\documentclass[11pt,a4paper]{article}
\usepackage[hyperref]{acl2019}
\usepackage{times}
\usepackage{latexsym}
\usepackage[TABBOTCAP]{subfigure}
\usepackage[shortlabels]{enumitem}

\usepackage{tikz-dependency}
\usepackage{times}
\usepackage{algorithm}
\usepackage{algpseudocode}
\usepackage{latexsym}
\usepackage{multirow}
\usepackage{booktabs}

\usepackage{color}
\usepackage{helvet}
\usepackage{textcomp}
\usepackage{booktabs}
\usepackage{multirow}
\usepackage{booktabs}
\usepackage{color}
\usepackage{graphicx}
\graphicspath{ {images/} }
\usepackage{amsmath}

\usepackage{hyperref}
\usepackage{url}

\usepackage{url}

\aclfinalcopy % Uncomment this line for the final submission
\def\aclpaperid{638} %  Enter the acl Paper ID here

\title{SP-10K: A Large-scale Evaluation Set for \\ Selectional Preference Acquisition}

\author{Hongming ZHANG, Hantian DING, and Yangqiu SONG \\
        Department of CSE, HKUST\\
        hzhangal@cse.ust.hk, hdingab@connect.ust.hk, yqsong@cse.ust.hk}
\date{}

\begin{document}
\maketitle
\begin{abstract}
% Selectional Preference (SP) is a commonly observed language phenomenon and plays a crucial role in a number of natural language processing tasks.
% % However, current SP models are rather limited due the shortcomings of existing evaluation methods.
% To provide a better evaluation method for SP models, we introduce SP-10K, a large-scale evaluation set that provides human ratings for the plausibility of 10,000 SP pairs over five SP relations, covering the 2,500 most frequent verbs, nouns, and adjectives in American English. 
% We then evaluate the performance of six representative SP acquisition methods on our dataset.
% To demonstrate the importance of our dataset, we also investigated the relationship between our collected SP pairs and the commonsense knowledge from ConceptNet5 and show the potential of using SP to represent commonsense knowledge.
% %To conclude, SP-10K provides \reviseyq{a broad coverage of evaluation} for selectional preference acquisition and reveals the relations between selectional preference and the underlying commonsense knowledge.

Selectional Preference (SP) is a commonly observed language phenomenon and proved to be useful in many natural language processing tasks.
To provide a better evaluation method for SP models, we introduce SP-10K, a large-scale evaluation set that provides human ratings for the plausibility of 10,000 SP pairs over five SP relations, covering 2,500 most frequent verbs, nouns, and adjectives in American English.
Three representative SP acquisition methods  based on pseudo-disambiguation are evaluated with SP-10K.
To demonstrate the importance of our dataset, we investigate the relationship between SP-10K and the commonsense knowledge in ConceptNet5 and show the potential of using SP to represent the commonsense knowledge.
We also use the Winograd Schema Challenge to prove that the proposed new SP relations are essential for the hard pronoun coreference resolution problem.

\end{abstract}

\section{Introduction}\label{sec-introduction}
% 1 page

% What is SP
Selectional Preference (SP) is a common phenomenon in human language that has been shown to be related to semantics~\cite{wilks1975preferential}. 
Here by SP we mean that, given a word and a dependency relation, human beings have preferences for which words are likely to be connected.
For instance, when seeing the verb `sing', it is highly plausible that its object is `a song', and when seeing the noun `air', it is highly plausible that its modifier is `fresh'.
%SP is important because it implicitly encodes information that is crucial for natural language understanding. 

% Where can we use SP
SP has been shown to be useful over a variety of tasks including sense disambiguation~\cite{resnik1997selectional}, semantic role classification~\cite{semantic_role_classification}, coreference clustering~\cite{hobbs1978resolving, DBLP:conf/coling/InoueMOOI16, DBLP:conf/emnlp/HeinzerlingMS17}, and machine translation~\cite{DBLP:conf/coling/TangXZG16}.
Given the importance of SP, the automatic acquisition of SP has become a well-known research subject in the NLP community.
However, current SP acquisition models are limited based on existing evaluation methods. We discuss two broadly used evaluation methods, human-labeled evaluation sets and the pseudo-disambiguation task. 

\begin{table}[t]
\footnotesize
\centering
\begin{tabular}{c|ccc}
	\toprule
    SP Evaluation Set & \#R & \#W & \#P  \\
    \midrule
    \cite{mcrae1998modeling} & 2 & 641 & 821  \\
    \cite{DBLP:journals/coling/KellerL03} & 3 & 571   & 540   \\
    \cite{pado2006combining}   & 3 & 180   & 207   \\
    \midrule
    SP-10K               & 5 & 2.5K & 10K  \\
    \bottomrule
	\end{tabular}
	\caption{  Statistics of Human-labeled SP Evaluation Sets. \#R, \#W, and \#P indicate the number of SP relation types, words, and pairs, respectively. }
%	\vspace{-0.15in}
	\label{tab:stat}

\end{table}

First, the most straightforward way to evaluate SP models is by asking human annotators. \cite{mcrae1998modeling}, \cite{DBLP:journals/coling/KellerL03}, and \cite{pado2006combining} proposed human-labeled SP evaluation sets containing hundreds of SP pairs (numbers are shown in Table~\ref{tab:stat}). However, these datasets are too small to cover the diversity of the SP task adequately. Moreover, they only considered one-hop relations, such as `verb-object' and `modifier-noun' pairs.
Aside from these relations, we believe that higher-order dependency relations may also reflect meaningful commonsense knowledge.
Consider the following two examples of hard pronoun resolution problems from the Winograd Schema Challenge~\cite{levesque2011winograd}:
\begin{itemize}

\setlength\itemsep{0.1em}
  \item (A) The fish ate the worm. It was hungry.
  \item (B) The fish ate the worm. It was tasty.
\end{itemize}

In (A), we can resolve `it' to `the fish' because it is more plausible that the subject of the verb `eat' is hungry.
On the other hand, for (B), we can resolve `it' to `the worm' because it is more likely that the object of the verb `eat' is tasty.
The above examples reflect the preferences between two two-hop dependency relations: `verb-object-modifier' and `verb-subject-modifier', which have not been investigated in previous works.

Second, pseudo-disambiguation has been a popular alternative evaluation method for the SP acquisition task~\cite{DBLP:conf/acl/RitterME10, DBLP:conf/emnlp/Cruys14}.
This way of SP acquisition trains a model based on pairs from a training corpus as positive examples and randomly generates fake pairs as negative examples, and then evaluates the model based on its ability on a test corpus by constructing positive and negative examples in the same way.
However, the pseudo-disambiguation task only evaluates how well a model fits the data, which could be biased.
The problem is that changing the corpus of training and testing may result in different conclusions.
Thus, it is less robust than collecting SP pairs by asking expert annotators as \cite{mcrae1998modeling}, \cite{DBLP:journals/coling/KellerL03}, and \cite{pado2006combining}, or even asking many ordinary people to vote for a commonsense agreement.
%This is especially problematic over corpora with noisy and ambiguous SP pairs, limiting the usefulness of learned models evaluated against pseudo-disambiguation for downstream tasks.
%However, pseudo-disambiguation can be easily influenced by the evaluation corpus and \reviseyq{cannot} represent the ground truth of SP.

The problems of these methods motivate the creation of a large-scale human-labeled SP evaluation set based on crowdsourcing, which can be used as the ground truth for the SP acquisition task.

% \begin{table}[t]
% \small
% \centering
% \begin{tabular}{c|c|c}
% 	\toprule
%     Relation & Head & Dependent \\
%     \midrule
%     `dobj' & verb & object \\
%     `nsubj' & verb & subject \\
%     `amod' & noun & modifier \\
%     `dobj\_amod' & verb & modifier of the object \\
%     `nsubj\_amod' & verb & modifier of the subject\\
%     \bottomrule
% 	\end{tabular}
%     \vspace{-0.05in}
% 	\caption{  SP relation definitions.} \label{tab:definition}
% 	\vspace{-0.15in}
% \end{table}
In this paper, we present SP-10K, which is unprecedented in both size and the number of SP relations. It contains 10,000 selectional triplets consisting of 2,500 frequent verbs, nouns, and adjectives in American English. Besides commonly used one-hop SP relations (`dobj', `nsubj', and `amod'), we introduce two novel two-hop SP relations (`dobj\_amod' and `nsubj\_amod').  
We first evaluate three representative SP acquisition methods using SP-10K and compare the capacity of the state-of-the-art pseudo-disambiguation approaches.
We then show the relationship between SP-10K and commonsense knowledge using ConceptNet5~\cite{DBLP:conf/lrec/SpeerH12} to demonstrate the potential of using SP to represent commonsense knowledge.
Finally, we use a subset of the Winograd Schema Challenge~\cite{levesque2011winograd} to prove that the proposed two-hop SP relations are essential for the hard pronoun coreference resolution.
SP-10K is available at: \url{https://github.com/HKUST-KnowComp/SP-10K}.

\section{Design of SP-10K}\label{sec-sp10k}

As discussed in~\cite{DBLP:journals/coling/HillRK15}, a high-quality evaluation resource should be: (1) clearly defined; (2) representative; and (3) consistent and reliable.

First, similar to existing human-labeled SP evaluation sets~\cite{mcrae1998modeling, DBLP:journals/coling/KellerL03, pado2006combining}, SP-10K uses the plausibility of selectional pairs as the annotation. Hence, SP-10K is clearly defined.
Second, compared to these existing evaluation sets, as shown in Table~\ref{tab:stat}, SP-10K covers a larger number of relations and SP pairs, making it a more representative evaluation set.
Finally, as discussed in Section~\ref{sec:interAgreement}, the annotation of SP-10K is consistent and reliable.

\subsection{Selectional Relations}

Traditionally, the study of SP has focused on three selectional relations: verb-subject, verb-object, and noun-adjective. 
%However, we believe that other selectional relations may also reflect meaningful commonsense knowledge. 
As demonstrated in Section~\ref{sec-introduction}, some verbs have a preference for the properties of their subjects and objects. For example, it is plausible to say that the subject of `eat' is hungry and the object of `eat' is tasty, but not the other way round. To capture such preferences, we propose two novel two-hop dependency relations, `dobj\_amod' and `nsubj\_amod'. Examples of these relations are presented in Table~\ref{tab:example}. In total, SP-10K contains five SP relations.

Following previous approaches~\cite{mcrae1998modeling, pado2006combining}, for the `dobj' and `nsubj' relations, we take a verb as the head and a noun as the dependent. Similarly, for `dobj\_amod' and `nsubj\_amod' relations, we take a verb as the head and an adjective as the dependent. Moreover, for the `amod' relation, we take a noun as the head and an adjective as the dependent.

\subsection{Candidate SP Pairs}
% \subsection{Vocabulary}

The selected vocabulary consists of 2,500 verbs, nouns, and adjectives from the 5,000 most frequent words\footnote{https://www.wordfrequency.info/free.asp} in the Corpus of Contemporary American English. 
%We believe that this corpus is representative, given that~\citet{francis1982frequency} and~\citet{solovyev2017dynamics} show that these words cover more than 80\% of daily used vocabulary. Moreover, these words are easy to understand, which can aid the annotation process.

\begin{table}[t]
    \small
		\centering
		\begin{tabular}{c|c|c}
			\toprule 
			Relation & Frequent & Random \\ 
			\midrule
			\multirow{ 2}{*}{`dobj'}  & (ask, question) & (ask, voting)\\
            						  & (ask, time)     & (ask, stability)\\
            \midrule
			\multirow{ 2}{*}{`nsubj'} & (people, eat)   & (textbook, eat)\\
            						  & (husband, eat)  & (stream, eat)\\
            \midrule
            \multirow{ 2}{*}{`amod'}  & (fresh, air)    & (rational, air)\\
            						  & (cold, air)     & (original, air)\\
            \midrule
            \multirow{ 2}{*}{`dobj\_amod'}  & (design, new)    & (design, official)\\
            						  & (design, original)     & (design, civil)\\
            \midrule
            \multirow{ 2}{*}{`nsubj\_amod'}  & (friendly, smile)    & (young, smile)\\
            						  & (symbolic, smile)     & (civilian, smile)\\
			\bottomrule
		\end{tabular}
		\caption{ Examples of candidate pairs for annotation. For the ease of understanding, the order of head and dependent may be different for various relations.} \label{tab:example}
	\end{table}
% \reviseyq{(now I can understand. Better to put an example for some of the pairs in this table..)}
For each SP relation, we provide two types of SP pairs for our annotators to label: frequent pairs and random pairs.
For each selectional relation, we first select the 500 most frequent heads. We then match each head with its two most frequently-paired dependents, as well as two randomly selected dependents from our vocabulary. As such, we retrieve 2,000 pairs for each relation.
Altogether, we retrieve 10,000 pairs for five selectional relations. These pairs are composed of 500 verbs, 1,343 nouns, and 657 adjectives. 
Examples of sampled pairs are presented in Table~\ref{tab:example}. 

\section{Annotation of SP Pairs}\label{sec-design}

% 3 pages 
We employ the Amazon Mechanical Turk platform (MTurk) for our annotations.\footnote{According to ~\cite{peer2017beyond}, Amazon MTurk (https://www.mturk.com/) has the largest worker population and highest annotation quality compared to other crowdsourcing services.} 

\subsection{Survey Design}

Following the SimLex-999 annotation guidelines~\cite{DBLP:journals/coling/HillRK15}, we invite at least 11 annotators to score each SP pair. We divide our 10,000 pairs into 100 surveys. Each survey contains 103 questions, three of which are checkpoint questions selected from the examples to control the labeling quality. Within a survey, all the questions are derived from the same selectional relation to improve the efficiency of survey completion.

Each survey\footnote{A sample survey is provided in the appendix.} consists of three parts. We begin by explaining the task to the annotators, including how to deal with the special case like multi-word expressions. Then, we present three examples to help the annotators better understand the task. Finally, we ask questions using the following templates (VERB, ADJ, and NOUN are place holders and will be replaced with the corresponding heads and dependents in the actual surveys.):

\begin{itemize}[leftmargin=*]
  \setlength\itemsep{0.1em}
  \item \textbf{dobj:} How suitable do you think it is if we use NOUN as the object of the verb VERB?
  \item \textbf{nsubj:} How suitable do you think it is if we use NOUN as the subject of the verb VERB?
  \item \textbf{amod:} How suitable do you think it is if we use ADJ to describe the noun NOUN?
  \item \textbf{dobj\_amod:} How suitable do you think it is if we use ADJ to describe the object of the verb VERB?
  \item \textbf{nsubj\_amod:} How suitable do you think it is if we use ADJ to describe the subject of the verb VERB?
\end{itemize}

For each question, the annotator is asked to select one of the following options: Perfectly match (5), Make sense (4), Normal (3), Seems weird (2), It's not applicable at all (1). 
We randomize the order of frequent and random pairs to prevent annotators from simply memorizing the question order.

% \subsection{Checkpoint Questions}
% To examine the quality of the annotation, We randomly insert the example questions in the instruction as the checkpoint questions into the survey. 

\begin{figure}[t]
	\centering
	\includegraphics[width=0.9\linewidth]{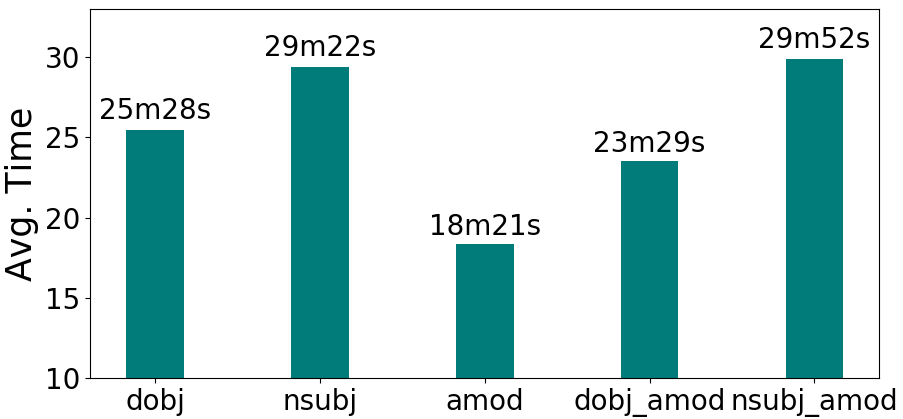}
    \caption{Average annotation time per 100 questions. `m' indicates minutes and `s' indicates seconds.} 
	\label{fig:Averagetime}
\end{figure}

\begin{table*}[t]
\small
	\centering
    \subtable[dobj]{
      \begin{tabular}{c|c}
            \toprule 
               SP  Pair & Plausibility \\
            \midrule
              (eat, meal)  & 10.00 \\
              (close, door) & 8.50 \\
              (convince, people) & 7.75 \\
              (touch, food) & 5.50 \\
              (hate, investment) & 4.00 \\
              (confront, impulse) & 2.78 \\

              (eat, mail) & 0.00 \\
                \bottomrule
            \end{tabular}
    }
    \subtable[nsubj]{
      \begin{tabular}{c|c}
            \toprule 
               SP  Pair & Plausibility \\
            \midrule
              (singer, sing)  & 10.00 \\
              (law, permit) & 7.78 \\
              (women, pray) & 5.83 \\
              (realm, remain) & 3.06 \\
              (victim, contain) & 2.22 \\
              (bar, act) & 1.39 \\
              (textbook, eat) & 0.00 \\

                \bottomrule

            \end{tabular}
    }
    \subtable[amod]{
      \begin{tabular}{c|c}
            \toprule 
              SP   Pair & Plausibility \\
            \midrule
              (fresh, air)  & 9.77 \\
              (new, method) & 8.89 \\
              (young, people) & 6.82 \\
              (medium, number) & 4.09\\
              (immediate, food) & 2.50\\
              (eager, price) & 1.36 \\
              (secret, wind) & 0.75 \\
                \bottomrule
            \end{tabular}
    }
    \subtable[dobj\_amod]{
      \begin{tabular}{c|c}
            \toprule 
              SP   Pair & Plausibility \\
            \midrule
              (lift, heavy \textit{object})  & 9.17 \\
              (design, new \textit{object}) & 8.00 \\
              (recall, previous \textit{object})  & 7.05 \\
              (attack, small \textit{object}) & 5.23 \\
              (drag, drunk \textit{object}) & 4.25 \\
              (inform, weird \textit{object}) & 3.64 \\
              (earn, rubber \textit{object})  & 0.63 \\
                \bottomrule
            \end{tabular}
    }
    \subtable[nsubj\_amod]{
      \begin{tabular}{c|c}
            \toprule 
              SP   Pair & Plausibility \\
            \midrule
              (friendly \textit{subject}, smile)  & 10.00 \\
              (evil \textit{subject}, attack) & 9.00 \\
              (recent \textit{subject}, demonstrate) & 6.00\\
              (random \textit{subject}, bear) & 4.00\\
              (happy \textit{subject}, steal)  & 2.25 \\
              (stable \textit{subject}, understand) & 1.75 \\
              (sunny \textit{subject}, make)  & 0.56 \\
                \bottomrule
            \end{tabular}
    }
%    \vspace{-0.1in}
	\caption{Sampled SP pairs from SP-10K and their plausibility ratings. \textit{object} and \textit{subject} are place holders to help understand the two-hop SP relations.} \label{tab:SampleResult}
%    \vspace{-0.2in}
\end{table*}

\subsection{Participants and Annotation}
We require that our annotators are `Master Workers', indicating reliable annotation records\footnote{ https://www.mturk.com/worker/help}, and that our annotators are either native English speakers or currently live and/or work in English-speaking locales. Based on these criteria, we identified 125 valid annotators. These annotators produced 130,575 ratings for a total cost of USD1,182.80. We support the multiple participation of annotators by ensuring that subsequent surveys are generated with their previously-unanswered questions.

From our annotation statistics, we notice that different selectional relations take different time to annotate. As shown in Figure~\ref{fig:Averagetime}, the annotators spent the least time on the `amod' relation, suggesting that the modifying relation is relatively easy to understand and judge. Another interesting finding is that the annotators spend more time on relations involving subjects than those involving objects, which is consistent with the observation proposed by~\cite{jackendoff1992semantic} that verbs have clearer preferences for objects than subjects.

\subsection{Post-processing}
We excluded ratings from annotators who (1) provided incorrect answers to any of the checkpoint questions or (2) demonstrated suspicious annotation patterns (e.g., marking all pairs as `normal').
After excluding based on this criteria, we obtained 100,532 valid annotations with an overall acceptance rate of 77\%.
% While performing post-processing, we noticed that some heads, e.g., `sing' or `eat', have very strong and clear preferences for their dependents. Annotators provided consistent and confident annotations for these head-dependents pairs.
% On the other hand, heads exhibiting weaker preferences, e.g., `want' or `get', were less consistently annotated. This is consistent with our understanding of SP.
% To ensure the quality of our evaluation set, for each relation, we select the top 1,000 consistently annotated pairs to form the \textsc{} subset.
% we label each of the 2,000 pairs for each SP relation as either `\textsc{consistent}' or `\textsc{ambiguous}' based on its labeling variance. Each of both labels \reviseyq{contains} 1,000 pairs for each relation.
% We provide pairs from both labels as part of the SP-10K dataset, but we only use the consistent pairs as our evaluation set in this paper.
% We only use the \textsc{strong} subset as our evaluation set in this paper.
We calculate the plausibility for each SP pair by taking the average rating for the pair over all (at least 10) valid annotations, then linearly scaling this average from the 1-5 to 0-10 interval. This approach is similar to the post-processing in~\cite{DBLP:journals/coling/HillRK15}. We present a sample of SP pairs in Table~\ref{tab:SampleResult}. 
Some of the pairs are interesting.
For example, for the dobj\_amod relation, annotators agree that lifting a heavy object is a usually used expression, while earning a rubber object is rare.
%We generally find that the pairs and their ratings are fairly reasonable.
% From the table we can see that, all of the pairs and their ratings \revisevk{are reasonable} to English speakers.

% \begin{table}[t]
% 	\small
% 	\centering
% 	\begin{tabular}{c|cc}
% 		\toprule 
% 		Eval set	 & IAA-1 & IAA-2 \\
% 		\midrule
%         WS-353 & 0.61 & - \\
%         WSIM-203 & 0.67 & 0.65 \\
%         Simlex-999 & 0.67 & 0.78 \\
%         SimVerb-3500 & 0.84 & 0.86\\
%         \midrule
% 		SP-10K (\textsc{strong}) (5,000) & 0.69 & 0.78\\
% %         SP-10K (\textsc{ambiguous}) (5,000) & 0.41 & 0.57\\
%         SP-10K (\textsc{all}) (10,000) & 0.56 & 0.70\\

% 			\bottomrule
% 		\end{tabular}
% 		\caption{Overall Inner-Annotator Agreement comparison. } \label{tab:interAgreement}
%        \vspace{-0.2in}
% \end{table}

\begin{table}[t]
	\small
	\centering
	\begin{tabular}{c|ccccc|c}
		\toprule 
			  & dobj & nsubj & amod & d\_a & n\_a & overall \\
		\midrule
        IAA  & 0.83 & 0.77 & 0.81 & 0.71 & 0.63 & 0.75 \\

			\bottomrule
		\end{tabular}
		\caption{Overall Inter-Annotator Agreement (IAA) of SP-10K. `d$\_$a' stands for dobj$\_$amod and `n$\_$a' stands for nsubj$\_$amod. } \label{tab:interAgreement}
\end{table}

\subsection{Inner-Annotator Agreement}\label{sec:interAgreement}

Following standard practices from previous datasets WSIM-203~\cite{DBLP:conf/emnlp/ReisingerM10} and Simlex-999~\cite{DBLP:journals/coling/HillRK15}, we employ Inter-Annotator Agreement (IAA), which computes the average correlation of an annotator with the average of all the other annotators, to evaluate the overall annotation quality.
As presented in Table~\ref{tab:interAgreement}, the overall IAA of SP-10K is $\rho$ = 0.75, which is comparable to existing datasets WSIM-203 (0.65) and Simlex-999 (0.78). 

Unsurprisingly, the IAA is not uniform across different SP relations. As shown in Table~\ref{tab:interAgreement}, complicated two-hop SP relations are more challenging and achieve relatively lower correlations than the simpler one-hop relations. This experimental result shows that two-hop relations are more difficult than one-hop SP relations.
We also notice that the agreements among annotators for SP relations involving the subjects of verbs are relatively low. 
The above observations are consistent with our earlier discussion on annotation time, and further support the claim that verbs have stronger preferences for their objects than their subjects.

\section{Evaluation of SP Acquisition Methods}\label{sec-eval}
% In this section, we evaluate representative SP acquisition methods on SP-10K.

To show the performance of existing SP acquisition methods and demonstrate the effect of different training corpora, we evaluate representative SP acquisition methods on SP-10K with following training corpora:
% \subsection{Corpus}

% We first introduce our training corpora.

\textbf{(1) Wiki:} Wikipedia is the largest free knowledge dataset. For this experiment, we select the English version of Wikipedia\footnote{https://dumps.wikimedia.org/enwiki/} and filter out pages containing fewer than 100 tokens and fewer than five hyperlinks. After filtering, our dataset contains over three million Wikipedia pages.

\textbf{(2) Yelp:} Yelp is a social media platform where users can write reviews for businesses, e.g., restaurants, hotels, etc. The latest release of the Yelp dataset\footnote{https://www.yelp.com/dataset/challenge} contains over five million reviews. 

\textbf{(3) New York Times (NYT):} The NYT~\cite{nyt} dataset contains over 1.8 million news articles from the NYT throughout 20 years (1987 - 2007).

We parsed these raw corpora using the Stanford dependency parser~\cite{DBLP:conf/lrec/SchusterM16}. Detailed statistics are shown in Table~\ref{tab:corpus}.

\begin{table}[t]
	\centering
    \small
	\begin{tabular}{c|ccc}
		\toprule 
			           & Wiki & Yelp & NYT \\
		\midrule
        \#(sentence)     & 82m  & 41m  & 56m \\
		\#(dobj pairs)         & 69m  & 33m  & 49m \\
		\#(nsubj pairs)        & 97m  & 70m  & 86m \\
        \#(amod pairs)        & 119m & 31m  & 65m \\
        \#(dobj\_amod pairs)   & 21m  & 8.1m & 14m \\
		\#(nsubj\_amod pairs)  & 16m  & 4.8m & 12m \\
		\bottomrule
		\end{tabular}
		\caption{ Training corpus statistics. `m' means millions. } \label{tab:corpus}
\end{table}

\begin{table*}[h]
	\centering
	\small
    \subtable[dobj]{
      \begin{tabular}{c|ccc}
            \toprule 
              Model   & Wiki & Yelp & NYT  \\
            \midrule
              PP  & 0.74$^{\star \dag}$ & \textbf{0.76}$^{\star \dag}$ & 0.74$^{\star}$ \\
              DS & 0.65 & 0.55& 0.63 \\
              NN & 0.68 & 0.55& 0.71 \\
            \bottomrule
            \end{tabular}
    }
    \subtable[nsubj]{
      \begin{tabular}{c|ccc}
            \toprule 
              Model   & Wiki & Yelp & NYT  \\
            \midrule
              PP  & \textbf{0.75}$^{\star }$ & 0.66$^{\star \dag}$ & 0.73$^{\star \dag}$ \\
              DS  & 0.59 & 0.46& 0.59 \\
              NN  & 0.70 & 0.54& 0.69 \\
            \bottomrule
            \end{tabular}
    }
    \subtable[amod]{
      \begin{tabular}{c|ccc}
            \toprule 
              Model   & Wiki & Yelp & NYT  \\
            \midrule
              PP  & \textbf{0.75}$^{\star \dag}$ & 0.71$^{\star \dag}$& 0.74$^{\star \dag}$ \\
              DS & 0.67 & 0.47& 0.62 \\
              NN & 0.68 & 0.50& 0.69 \\
            \bottomrule
            \end{tabular}
    }
    \subtable[dobj\_amod]{
      \begin{tabular}{c|ccc}
            \toprule 
              Model   & Wiki & Yelp & NYT  \\
            \midrule
              PP  & \textbf{0.65}$^{\star}$ & 0.62$^{\star \dag}$& 0.63$^{\star}$ \\
              DS & 0.55 & 0.47& 0.55 \\
              NN & 0.62 & 0.52& 0.64 \\
            \bottomrule
            \end{tabular}
    }
    \subtable[nsubj\_amod]{
      \begin{tabular}{c|ccc}
            \toprule 
              Model   & Wiki & Yelp & NYT  \\
            \midrule
              PP  & 0.52$^{\star \dag}$ & 0.36& \textbf{0.54}$^{\star \dag}$ \\
              DS & 0.46 & 0.33& 0.47 \\
              NN & 0.46 & 0.32& 0.47 \\
            \bottomrule
            \end{tabular}
    }
    \subtable[overall]{
      \begin{tabular}{c|ccc}
            \toprule 
              Model   & Wiki & Yelp & NYT  \\
            \midrule
              PP  & \textbf{0.68}$^{\star \dag}$ & 0.62$^{\star \dag}$& \textbf{0.68}$^{\star \dag}$ \\
              DS  & 0.58 & 0.46& 0.57 \\
              NN  & 0.63 & 0.49& 0.64 \\
            \bottomrule
            \end{tabular}
    }
	\caption{ Performance of different corpora and methods on SP-10K. Average Spearman $\rho$ scores are reported. $\star$ indicates statistical significant (p \textless 0.005) over DS and $\dag$ indicates statistical significant (p \textless 0.005) over NN. For each SP relation, rows represent different acquisition methods and columns represent different corpora. The best performed model for each relation is annotated with \textbf{bold} font.} \label{tab:Performance}
\end{table*}
\subsection{Methods}
We now introduce SP acquisition methods.

\textbf{Posterior Probability (PP): }~\cite{resnik1997selectional} proposes PP as a means of acquiring SP knowledge from raw corpora. Given a head $h$, a relation $r$, and a dependent $d$, PP uses the following probability to predict the plausibility:
\begin{equation}\label{naturalprobability}
			P_r(d|h) = \frac{C_r(h, d)}{C_r(h)}, 
		\end{equation}
where $C_r(h)$ and $C_r(h, d)$ mean how many times $p$ and the head-dependent pair ($h$, $d$) appear in the relation $r$ respectively.

\textbf{Distributional Similarity (DS): } \cite{DBLP:journals/coling/ErkPP10} describes a method that uses corpus-driven DS metrics for the induction of SP. Given a head $h$, a relation $r$, and a dependent $d$, DS uses the following equation to predict the plausibility:
\begin{equation}\label{DS}
	S(h, r, d) = \sum_{d^\prime \in O_{r,h}}\frac{w(d, d^\prime)}{Z_{r,h}} \cdot s(d, d^\prime),
\end{equation}
where $O_{r,h}$ is the set of dependents that have been attested with head $p$ and relation $r$, $w(d, d^\prime)$ is the weight function, and $Z_{r,h}$ is the normalization factor. 
We use the frequency of a pair of $(h, d^\prime)$ as the weighting function and the cosine similarity of their GloVe embedding~\cite{DBLP:conf/emnlp/PenningtonSM14} as the similarity function $s(d, d^\prime)$, given the relative popularity of these embeddings.

\textbf{Neural Network (NN): } \cite{DBLP:conf/emnlp/Cruys14} proposes a NN-based method for the SP acquisition task.
The main framework is a two-layer fully-connected NN.
For each SP pair ($h$, $d$), the framework uses the concatenation of embeddings [${\bf v}_h$, ${\bf v}_d$] as the input to the NN, where ${\bf v}_h$, ${\bf v}_d$ are randomly initialized word embeddings for words $h$ and $d$ respectively.
The ranking-loss~\cite{DBLP:conf/icml/CollobertW08} is used as the training objective, where positive examples consist of all the SP pairs in the corpus and negative examples are randomly generated.
During the training process, both model parameters and embeddings are jointly updated.
We use the original paper's experimental setting to conduct our experiment.

\subsection{Results and Analysis}
We report the average Spearman $\rho$ in Table~\ref{tab:Performance} as our performance measure. We have following interesting observations.

(1) Choice of training corpus can influence the SP acquisition models. For the same method, the general corpora, i.e., Wiki and NYT, outperform the domain specific corpus, i.e., Yelp.
Yelp performs best on the `dobj' relation and comparably on the `dobj\_amod' relation, which indicates the language use on Yelp may better reflect the plausibility of objects rather than of subjects.

(2) As reported by \cite{DBLP:conf/emnlp/Cruys14}, the NN-based method performs very well on the pseudo-disambiguation task.
However, this method has limited effectiveness on our dataset, which shows that pseudo-disambiguation cannot effectively represent ground truth SP. This further demonstrates the value of SP-10K as an evaluation set of SP acquisition. 

(3) The overall performance of existing methods is quite lackluster, suggesting that these models insufficiently address the SP acquisition task. We hope that the release of our dataset will motivate efforts at deriving knowledge from SP and exploring the SP acquisition task.

\section{SP and Commonsense Knowledge}\label{sec-commonsense}

In this section, we quantitatively analyze the relationship between SP and commonsense knowledge.
Currently, the largest commonsense knowledge dataset is the Open Mind Common Sense (OMCS) from the ConceptNet 5~\cite{DBLP:conf/lrec/SpeerH12} knowledge base. The OMCS contains 600k crowdsourced commonsense triplets such as (food, UsedFor, eat) and (wind, CapableOf, blow to east). 
All of the relations in OMCS are human-defined.
% Compared to the knowledge in the OMCS dataset, SP only relies on dependency information, which naturally exists in natural language, which decides that SP can be more easily acquired from corpora.
In comparison, SP only relies on naturally occurring dependency relations, which can be accurately identified using existing parsing tools~\cite{DBLP:conf/lrec/SchusterM16}.
%Furthermore, SP acquisition only relies on dependency information, whereas OMCS requires additional techniques such as knowledge graph embedding and matching. 

We aim to demonstrate how SP related to commonsense knowledge. Building relationships between SP and human-defined relations has two advantages: 
(1) We may be able to directly acquire commonsense knowledge through SP acquisition techniques.
(2) We may be able solve commonsense reasoning tasks from the perspective of SP, as illustrated through the two Winograd examples in Section~\ref{sec-introduction}.
These advantages motivate exploring the potential of using SP to represent commonsense knowledge.

% On the contrary, selectional preferences only rely on dependency relation, which can be precisely identified with current NLP tools like Stanford Enhanced++\cite{DBLP:conf/lrec/SchusterM16}. Compared with OMCS, SP knowledge can be more easily acquired and applied in other NLP tasks. 

\subsection{SP Pairs and OMCS Triplets}

% \revisevk{NOTE: Can we explain how the definition of SP relates to commonsense knowledge more explicitly?} 
% Based on the definition of SP, o
% Previous version: Our hypothesis is that more plausible SP pairs are closer to human commonsense knowledge. 
We hypothesize that the plausibility of an SP pair relates to how closely the pair aligns with human commonsense knowledge. As such, the more plausible pairs in SP-10K should be more likely to be covered by the OMCS dataset.

\begin{table}[t]
	\centering
    \small
	\begin{tabular}{c|ccc}
		\toprule 
			     Group       & \#Pairs & \#Exact Match & \#Partial Match \\
                     & & (Percentage) & (Percentage)\\
		\midrule
		Perfect     & 755   & 85 (11.26\%) & 287 (38.01\%) \\
        Good  & 2,600 & 67 (2.58\%)  & 885 (34.04\%) \\
        Normal      & 2,809 & 20 (0.71\%)  & 504 (17.94\%) \\
        Unusual     & 2,396 & 6  (0.25\%)  & 187 (7.80\%)  \\
        Impossible  & 1,440 & 5  (0.35\%)  & 82  (5.69\%)  \\
		\bottomrule
		\end{tabular}
		\caption{ Matching statistics of SP pairs by plausibility. } \label{tab:commonsense}
%        \vspace{-0.2in}
\end{table}
% 1.5 page

\begin{figure*}[tb]
    \centering
	\subfigure[Exact Match]{\label{fig:exact}
		\includegraphics[width=0.48\linewidth]{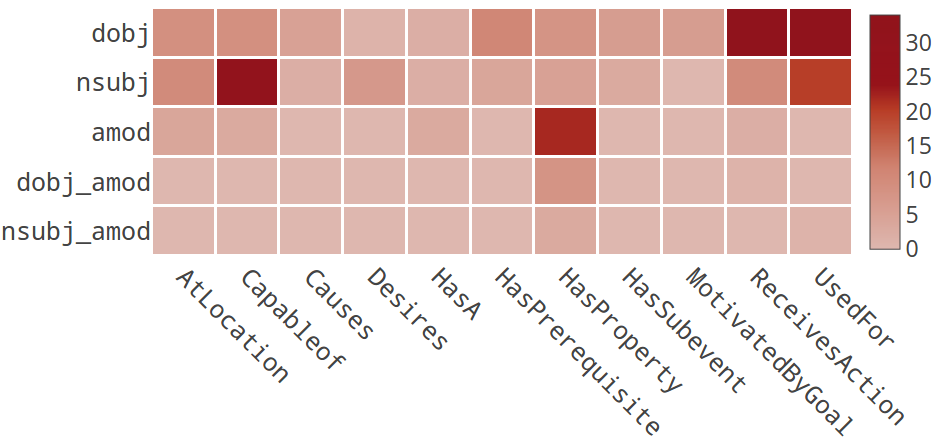}
	}        		
	\subfigure[Partial Match]{\label{fig:partial}
		\includegraphics[width=0.48\linewidth]{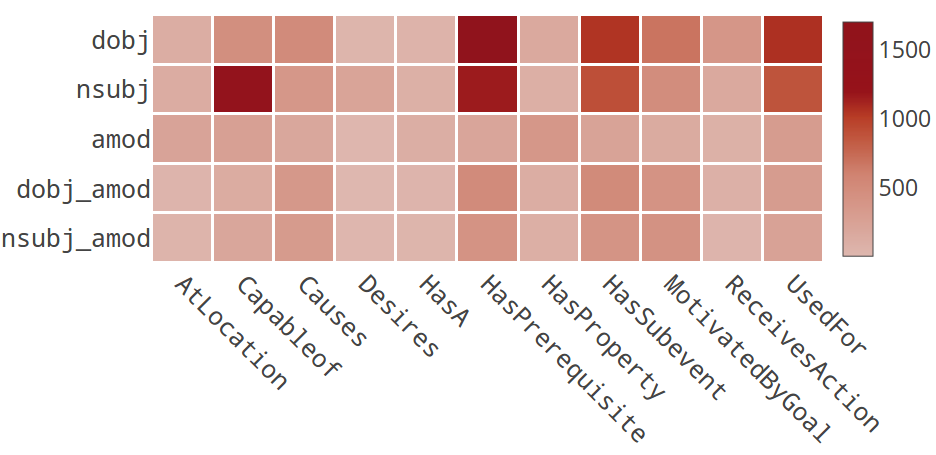}
	}        		  		
%    \vspace{-0.1in}
	\caption{Matched SP relations and OMCS relations. Interesting relation matches such as `dobj' versus `UserFor', `nsubj' versus `CapableOf', and `amod' versus `HasProperty' are observed.} 
	\label{fig:correspondingrelation}
\end{figure*}

\begin{table*}[t]
	\centering
    \small
    \subtable[Perfect group (Plausibility: 8-10)]{
    \begin{tabular}{c|c}
		\toprule 
			     SP relation & SP pair versus OMCS triplets \\
		\midrule
        \multirow{ 2}{*}{`dobj'} & (sing, song) (9.25/10)\\
        						& (song, UsedFor, sing)\\
        \midrule
        \multirow{ 2}{*}{`nsubj'} & (phone, ring) (8.75/10)\\
        						& (phone, CapableOf, ring)\\
        \midrule
        \multirow{ 2}{*}{`amod'} & (cold, water) (8.86/10)\\
        							& (water, HasProperty, cold)\\
        \midrule
        \multirow{ 2}{*}{`dobj\_amod'} & (create, new) (8.25/10)\\
        								& (create idea, UsedFor, invent new things)\\
        \midrule
        \multirow{ 2}{*}{`nsubj\_amod'} & (hungry, eat) (10.00/10)\\
        							& (eat, MotivatedByGoal, are hungry)\\
        
		\bottomrule
		\end{tabular}
	}
    \subtable[Impossible group (Plausibility: 0-2)]{
    \begin{tabular}{c|c}
		\toprule 
			     SP relation & SP pair versus OMCS triplets \\
		\midrule
        \multirow{ 2}{*}{`dobj'} & (eat, mail) (0.00/10)\\
        						& (mail letter, HasSubevent, eat cheese)\\
        \midrule
        \multirow{ 2}{*}{`nsubj'} & (library, love) (1.25/10)\\
        						& (love, Atlocation, library)\\
        \midrule
        \multirow{ 2}{*}{`amod'} & (red, child) (0.68/10)\\
        							& (child wagon, HasProperty, red)\\
        \midrule
        \multirow{ 2}{*}{`dobj\_amod'} & (drive, bottom) (1.50/10)\\
        								& (drive car, HasSubevent, bottom out)\\
        \midrule
        \multirow{ 2}{*}{`nsubj\_amod'} & (fun, hurt) (1.50/10)\\
        							& (having fun, HasSubevent, get hurt)\\
		\bottomrule
		\end{tabular}
	}
	\caption{ Examples of OMCS-covered SP pairs and their corresponding OMCS triplets.} \label{tab:commonsensedemo}
\end{table*}

Using plausibility as our criterion, we split the 10,000 SP pairs into five groups: Perfect (8-10), Good (6-8), Normal (4-6), Unusual (2-4), and Impossible (0-2).
As OMCS triplets contain phrases and SP pairs only contain words, we use two methods to match SP pairs with OMCS triplets. (1) Exact Match: we identify triplets in OMCS where the two dependents are exactly the same as the two words in an SP pair. (2) Partial Match: we identify triplets in OMCS where the two dependents contain the two words in an SP pair.
We count SP pairs that fulfill either of these matching methods as covered by OMCS. Note that exact matches are not double-counted as partial matches.

As shown in Table~\ref{tab:commonsense}, almost 50\% of SP pairs in the perfect group are covered by OMCS. In contrast, only about 6\% of SP pairs from the impossible group are covered. 
More plausible selectional preference pairs are more likely to be covered by OMCS, which supports our hypothesis of more plausible SP pairs being more closely aligned with human commonsense knowledge.

\subsection{SP and Human-defined Relations}

To show the connection between SP relations and human-defined relations, we visualize all matching (SP pair, OMCS triplet) tuples in Figure~\ref{fig:correspondingrelation}. A darker color indicates a greater number of matched tuples, which in turn suggests a stronger connection between the two relations.
% to visualize the \revisevk{commonly} matched relations in OMCS, as shown in Figure~\ref{fig:correspondingrelation}.
% As shown in Figure~\ref{fig:correspondingrelation}, darker color means that more matchings between the two relations are identified.
% \revisevk{NOTE: By `all covered pairs', do you mean both exact and partial matches?}

We observe some clear and reasonable matches such as (`dobj', `UsedFor'), (`nsubj', `CapableOf'), and (`amod', `HasProperty'), 
which demonstrates that some simple human-defined relations like `UsedFor', `CapableOf', and `HasProperty' are related to corresponding SP relations.
% and are optimistic that SP acquisition models will be able to uncover many more such matches between SP and human-defined commonsense knowledge.
% Considering the size of SP-10K, it is normal that only a small portion of the OMCS triplets are covered by SP pairs in SP-10K for now.
% But if we can create a high quality SP models through SP acquisition, some of the above human-defined commonsense relation might be represented by the corresponding SP relations.
% Despite the currently limited coverage of OMCS triplets over the SP pairs in SP-10K, we are optimistic that SP acquisition models will be able to uncover many more matches between SP relations and human-defined commonsense relations.
We also notice that the five SP relations in SP-10K seldom match some OMCS relations such as `HasA' and `HasSubevent', which indicates a need for additional SP relations or even the combination of different SP relations. 
% For example, `HasA' represents a `noun-noun' preference and `HasSubevent' represents a `verb-verb' preference.
We leave it for our future work.
% This visualization shows that some of the human-defined commonsense knowledge can be effectively represented by corresponding SP relations and some others may need more SP relations.

\subsection{Case Study}

% We also select the covered pairs from the perfect and impossible groups for demonstration.\footnote{More examples are provided in the appendix.}
% For the perfect group, we find that human-defined commonsense triplets can be elegantly represented by corresponding SP pairs.
We present a selection of covered pairs from the perfect and impossible groups in Table~\ref{tab:commonsensedemo}\footnote{More examples are provided in the appendix.}. For the perfect group, we find that human-defined commonsense triplets often have neatly corresponding SP pairs.
On the other hand, for the impossible group, SP pairs are covered by OMCS either because of incidental overlap with a non-keyword, e.g., `child' in `child wagon', or because of the low quality of some OMCS triplets. This further illustrates that OMCS still has room for improvement and that SP may provide an effective way to improve commonsense knowledge.

\section{Importance of Multi-hop SP}\label{sec-wino}
As introduced in Section~\ref{sec-introduction}, one novel contribution of this paper is the two-hop Selectional Preference relations: `nsubj\_amod' and `dobj\_amod'. To demonstrate their effectiveness, we select a subset\footnote{We select all examples that use one adjective to describe the targeting pronoun and all selected questions are listed in the appendix.} of the Winograd Schema Challenge dataset~\cite{levesque2011winograd}, which leverages the two-hop selectional preference knowledge to solve. In total, we have 72 questions out of overall 285 questions.
The selected Winograd question is defined as follows: Given one sentence $s$ containing two candidates ($n_1$, $n_2$) and one pronoun $p$, which is described with one adjective $a$, we need to find which candidate is the pronoun referring to. One example is as follows:
\begin{itemize}
    \item \textit{Jim} yelled at \textit{Kevin} because \textbf{he} was so upset.
\end{itemize}
We need to correctly finds out \textbf{he} refers to \textit{Jim} rather than \textit{Kevin}.
These tasks are quite challenging as both the Stanford coreNLP coreference system and the current state-of-the-art end-to-end coreference model~\cite{lee2018higher} cannot solve them.
\begin{table}[t]
    \centering
    \small
    \begin{tabular}{c|ccc|cc}
        \toprule
        Model & Correct & Wrong & NA & $A_p$ & $A_o$ \\
        \midrule
        Stanford & 33 & 35 & 4 & 48.5\% & 48.6\% \\
        End2end & 36 & 36 & 0 & 50.0\% & 50.0\% \\
        \midrule
        PP & 36 & 19 & 17 & 65.5\% & \textbf{61.8\%} \\
        \midrule
        SP-10K & 13 & 0 & 59 & \textbf{100\%} & 59.0\% \\
        \bottomrule
         
    \end{tabular}
    \caption{Result of different models on the subset of Winograd Schema Challenge. $NA$ means that the model cannot give a prediction, $A_p$ means the accuracy of predict examples without $NA$ examples, and $A_o$ means the overall accuracy.}
    \label{tab:Wino}
\end{table}
To solve that problem from the perspective of selectional preference (SP), we first parse the sentence and get the dependency relations related to the two candidates. If they appear as the subject or the object of the verb $h$, we will then check the SP score of the head-dependent pair ($h$, $d$) on relations `nsubj\_amod' and `dobj\_amod' respectively.
After that, we compare the SP score of two candidates and select the higher one as the prediction result. 
If they have the same SP score, we will make no prediction.

We show the result of collected human-labeled data in `SP-10K' and the best-performed model, Posterior Probability (PP), trained with Wikipedia corpus in Table~\ref{tab:Wino}.
From the result, we can see that `SP-10K' can solve that problem with very high precision. 
But as we only label 4,000 multi-hop pairs, the overall coverage is limited.
On the other hand, automatic SP acquisition method PP can cover more questions, but the precision also drops due to the noise of the collected SP knowledge.
The experimental result shows that if we can automatically build a good multi-hop SP model, we could make some steps towards solving the hard pronoun coreference task, which is viewed a vital task of natural language understanding.
%  helps prove the value of the SP acquisition task and the proposed SP-10K dataset.

\section{Related Work}\label{sec-relatedWork}

% 1/2 page
% refer to word-level rather than synset-level selectional preference, which 
% Compared to synset-level SP, word-level SP does not require linguist annotation and can be more easily applied in NLP tasks.

% As a common phenomenon in human language, selectional preference can be used to deliver some hidden information, which is not shown in the sentence but could be crucial for human to understand. 
	% For instance, `I eat something', as human naturally understand the selectional preference of verb `eat', we can know decode this `something' to some kind of food and it may have some property like `tasty', `delicious', even though `something' itself has no semantics meaning at all.
% This kind of information can also be viewed as commonsense knowledge.

As one important language phenomenon, SP is  considered related to the Semantics Fit~\cite{mcrae1998modeling} and has been proved helpful in a series of downstream tasks including machine translation~\cite{DBLP:conf/coling/TangXZG16}, sense disambiguation~\cite{resnik1997selectional}, coreference resolution~\cite{hobbs1978resolving, DBLP:conf/coling/InoueMOOI16,DBLP:conf/icmlc2/ZhangS18}, and semantic role classification~\cite{semantic_role_classification}.

% For example, \citet{mechura2008selectional} demonstrates that despite minor differences, SP is a cross-lingual phenomenon, making it helpful for . Given that SP pertains to the human-interpreted hidden properties or semantic meanings behind words, it can also be used for related tasks such as 
% Moreover, existing works claim that SP is helpful for coreference resolution~\cite{DBLP:conf/coling/InoueMOOI16}, especially for the pronoun coreference resolution task~\cite{hobbs1978resolving}. The latter task is very challenging because pronouns themselves do not have any semantic meaning. 
%Altogether, these tasks show that SP is an important language phenomenon that can benefit the whole NLP community.

Several algorithms attempt to acquire SP automatically from raw corpora~\cite{resnik1997selectional,rooth1999inducing,DBLP:journals/coling/ErkPP10,santus2017measuring}.
However, \cite{mechura2008selectional} reveals that creating a high-quality SP model is difficult due to the noisiness and ambiguity of raw corpora. 
Several approaches attempt to address this issue by applying state-of-the-art word embeddings and neural networks to the automatic acquisition of SP~\cite{DBLP:conf/acl/LevyG14, DBLP:conf/emnlp/Cruys14}. Despite these efforts, the quality of learned SP models remains questionable due to the shortcomings of existing SP acquisition evaluation methods.

Currently, the most popular evaluation method for SP acquisition is the pseudo-disambiguation~\cite{DBLP:conf/acl/RitterME10, DBLP:conf/emnlp/Cruys14}. 
However, pseudo-disambiguation can be easily influenced by the aforementioned noisiness of evaluation corpora and cannot represent ground truth SP.
Experiments in this paper prove that the model performs well on the pseudo-disambiguation task may not correlate well with the human-labeled ground truth.
As for the ground truth, there are three human-labeled ground truth SP evaluation sets~\cite{mcrae1998modeling, DBLP:journals/coling/KellerL03, pado2006combining}. These evaluation sets score SP pairs based on their plausibility as determined by human evaluators. However, these datasets are of small sizes.
Compared to current evaluation methods, SP-10K is a human-annotated large-scale evaluation set and contains 10,000 SP pairs over five SP relations.

%2 pages
\section{Conclusion}\label{sec-conclusion}

% 1/4 page
In this work, we present SP-10K, a large-scale human-labeled evaluation set for selectional preference.
Compared with other evaluation methods, SP-10K has much larger coverage and can better represent ground truth SP.
Two novel two-hop SP relations `dobj\_amod' and `nsubj\_amod' are also introduced.
We evaluate three representative SP acquisition methods with our dataset.
After that, we demonstrate the potential of using SP to represent commonsense knowledge, which can be beneficial for the acquisition and application of commonsense knowledge.
In the end, we demonstrate the importance of the two-hop relations with a subset of the Winograd Schema Challenge.
% \reviseyq{(After reading the whole paper, I feel it is now a really interesting and solid one.)}
% In the future, we will explore additional SP relations and their applications to commonsense understanding tasks.
% We will release all the annotated data to motivate more further work on selectional preference.

\section*{Ackowledgment}
This paper was partially supported by the Early Career Scheme (ECS, No.26206717) from Research Grants Council in Hong Kong. In addition, Hongming Zhang has been supported by the Hong Kong Ph.D. Fellowship.
We also thank Intel Corporation for supporting our deep learning related research, and the anonymous reviewers for their valuable comments and suggestions that help improving the quality of this paper.
\bibliography{naaclhlt2019}
\bibliographystyle{acl_natbib}

\clearpage
\appendix

% 	\maketitle
\section{Appendix}
\subsection{Survey Example}
	\begin{figure*}[h]
		\centering
		\onecolumn\includegraphics[width=\linewidth]{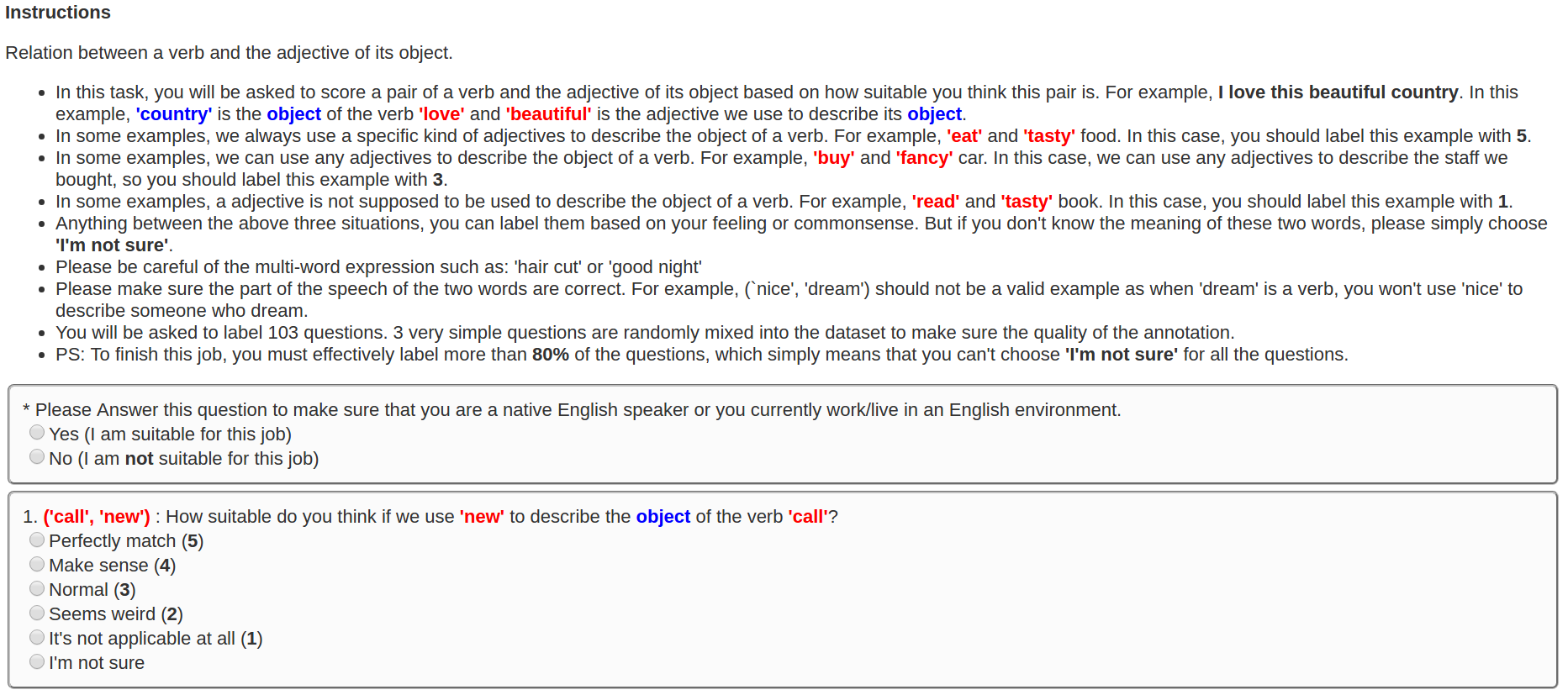}
		\caption{\normalsize Survey Example (As all the surveys are very similar to each other, we only show the survey for relation `dobj\_amod' as a demonstration).} 
		\label{fig:dobj_survey}
	\end{figure*}
\subsection{Selected Winograd Schema Challenge Questions}

The question id of selected 72 questions from the original 285 Winograd Schema Challenge\footnote{https://cs.nyu.edu/faculty/davise/papers/WinogradSchemas/WSCollection.xml} questions are as following: 3, 4, 7, 8, 15, 16, 19, 20, 35, 36, 39, 40, 43, 44, 45, 46, 51, 52, 71, 72, 73, 74, 75, 76, 77, 78, 79, 80, 87, 88, 89, 90, 97, 98, 107, 108, 109, 110, 111, 112, 119, 120, 131, 132, 147, 148, 150, 153, 154, 157, 158, 171, 172, 179, 180, 185, 186, 199, 200, 227, 228, 247, 248, 251, 252, 256, 257, 262, 263, 265, 282, 284.

	\clearpage

	\begin{table*}[t]
	\centering
	\small
	\subtable[\normalsize Perfect group in SP-10K]{
		\begin{tabular}{c|c}
			\toprule 
			SP relation & SP pair versus OMCS triplets \\
			\midrule
			\multirow{ 2}{*}{`dobj'} & (tell, story) (9.25/10)\\
			& (story, UsedFor, tell)\\
			\midrule
			\multirow{ 2}{*}{`nsubj'} & (author, write) (9.00/10)\\
			& (author, CapableOf, write)\\
			\midrule
			\multirow{ 2}{*}{`amod'} & (comfortable, chair) (8.25/10)\\
			& (chair, HasProperty, comfortable)\\
			\midrule
			\multirow{ 2}{*}{`dobj\_amod'} & (predict, future \textit{object}) (9.32/10)\\
			& (future, ReceivesAction, predict)\\
			\midrule
			\multirow{ 2}{*}{`nsubj\_amod'} & \multirow{ 2}{*}{-}\\
			&\\
			
			\bottomrule
		\end{tabular}
	}
		\subtable[\normalsize Good group in SP-10K]{
			\begin{tabular}{c|c}
				\toprule 
				SP relation & SP pair versus OMCS triplets \\
				\midrule
				\multirow{ 2}{*}{`dobj'} & (play, game) (7.75/10)\\
				& (game, UsedFor, play)\\
				\midrule
				\multirow{ 2}{*}{`nsubj'} & (water, pour) (7.75/10)\\
				& (water, CapableOf, pour)\\
				\midrule
				\multirow{ 2}{*}{`amod'} & (hard, rock) (7.75/10)\\
				& (rock, HasProperty, hard)\\
				\midrule
				\multirow{ 2}{*}{`dobj\_amod'} & (build, large \textit{object}) (7.00/10)\\
				& (build, HasProperty, large)\\
				\midrule
				\multirow{ 2}{*}{`nsubj\_amod'} & (illegal \textit{subject}, steal) (6.50/10)\\
				& (steal, HasProperty, illegal)\\
				
				\bottomrule
			\end{tabular}
		}
		\subtable[\normalsize Normal group in SP-10K]{
			\begin{tabular}{c|c}
				\toprule 
				SP relation & SP pair versus OMCS triplets \\
				\midrule
				\multirow{ 2}{*}{`dobj'} & (lock, key) (4.75/10)\\
				& (key, AtLocation, lock)\\
				\midrule
				\multirow{ 2}{*}{`nsubj'} &  (name, change) (5.50/10)\\
				& (name, CapableOf, change)\\
				\midrule
				\multirow{ 2}{*}{`amod'} & (high, seat) (5.25/10)\\
				& (seat, HasProperty, high)\\
				\midrule
				\multirow{ 2}{*}{`dobj\_amod'} &\multirow{ 2}{*}{-} \\
				& \\
				\midrule
				\multirow{ 2}{*}{`nsubj\_amod'} & (blue \textit{subject}, dress) (4.25/10)\\
				& (dress, HasProperty, blue)\\
				
				\bottomrule
			\end{tabular}
		}
		\subtable[\normalsize Unusual group in SP-10K]{
			\begin{tabular}{c|c}
				\toprule 
				SP relation & SP pair versus OMCS triplets \\
				\midrule
				\multirow{ 2}{*}{`dobj'} & (think, time) (2.75/10) \\
				& (time, UsedFor, think)\\
				\midrule
				\multirow{ 2}{*}{`nsubj'} & (restaurant, place) (3.75/10)\\
				& (restaurant, IsA, place)\\
				\midrule
				\multirow{ 2}{*}{`amod'} & \multirow{ 2}{*}{-}\\
				&\\
				\midrule
				\multirow{ 2}{*}{`dobj\_amod'} & \multirow{ 2}{*}{-}\\
				&\\
				\midrule
				\multirow{ 2}{*}{`nsubj\_amod'} & \multirow{ 2}{*}{-} \\
				& \\
				
				\bottomrule
			\end{tabular}
		}
	\subtable[\normalsize Impossible group in SP-10K]{
		\begin{tabular}{c|c}
			\toprule 
			SP relation & SP pair versus OMCS triplets \\
			\midrule
			\multirow{ 2}{*}{`dobj'} & (talk, wine) (1.75/10)\\
			& (wine, CauseDesire, talk)\\
			\midrule
			\multirow{ 2}{*}{`nsubj'} & (library, love) (1.25/10)\\
			& (love, Atlocation, library)\\
			\midrule
			\multirow{ 2}{*}{`amod'} & (high, teacher) (1.82/10)\\
			& (teacher, HasProperty, high)\\
			\midrule
			\multirow{ 2}{*}{`dobj\_amod'} & \multirow{ 2}{*}{-}\\
			&\\
			\midrule
			\multirow{ 2}{*}{`nsubj\_amod'} & \multirow{ 2}{*}{-}\\
			&\\
			\bottomrule
		\end{tabular}
	}
	\caption{\normalsize Examples of covered SP pairs and corresponding OMCS triplets (Exact Match).} 
\end{table*}

	\begin{table*}[h]
		\centering
		\small
		\subtable[\normalsize Perfect group in SP-10K]{
			\begin{tabular}{c|c}
				\toprule 
				SP relation & SP pair versus OMCS triplets \\
				\midrule
				\multirow{ 2}{*}{`dobj'} & (write, book) (8.25/10)\\
				& (book, ReceivesAction, write to be read)\\
				\midrule
				\multirow{ 2}{*}{`nsubj'} & (student, learn) (8.75/10)\\
				& (A student, CapableOf, learn calculus)\\
				\midrule
				\multirow{ 2}{*}{`amod'} & (long, story) (8.86/10)\\
				& (a story, HasProperty, long or short)\\
				\midrule
				\multirow{ 2}{*}{`dobj\_amod'} & (create, new \textit{object}) (8.25/10)\\
				& (create idea, UsedFor, invent new things)\\
				\midrule
				\multirow{ 2}{*}{`nsubj\_amod'} & (hungry \textit{subject}, eat) (10.00/10)\\
				& (eat, MotivatedByGoal, are hungry)\\
				
				\bottomrule
			\end{tabular}
		}
		\subtable[\normalsize Good group in SP-10K]{
			\begin{tabular}{c|c}
				\toprule 
				SP relation & SP pair versus OMCS triplets \\
				\midrule
				\multirow{ 2}{*}{`dobj'} & (take, time) (7.25/10)\\
				& (take bus, UsedFor, save time)\\
				\midrule
				\multirow{ 2}{*}{`nsubj'} & (people, use) (7.50/10)\\
				& (people, CapableOf, use money)\\
				\midrule
				\multirow{ 2}{*}{`amod'} & (new, house) (7.50/10)\\
				& (some house, HasProperty, new)\\
				\midrule
				\multirow{ 2}{*}{`dobj\_amod'} & (see, new \textit{object}) (6.50/10)\\
				& (see art, Causes, new ideas)\\
				\midrule
				\multirow{ 2}{*}{`nsubj\_amod'} & (patient \textit{subject}, wait) (7.50/10)\\
				& (wait in line, HasPrerequisite, be patient)\\
				
				\bottomrule
			\end{tabular}
		}
		\subtable[\normalsize Normal group in SP-10K]{
			\begin{tabular}{c|c}
				\toprule 
				SP relation & SP pair versus OMCS triplets \\
				\midrule
				\multirow{ 2}{*}{`dobj'} & (keep, work) (4.00/10) \\
				& (go to work, UsedFor, keep job)\\
				\midrule
				\multirow{ 2}{*}{`nsubj'} & (line, move) (4.72/10)\\
				& (line, IsA, move slowly)\\
				\midrule
				\multirow{ 2}{*}{`amod'} & (large, tree) (5.00/10)\\
				& (oak tree, IsA, large plant)\\
				\midrule
				\multirow{ 2}{*}{`dobj\_amod'} & (tell, good \textit{object}) (4.50/10)\\
				& (tell story, Causes, feel good)\\
				\midrule
				\multirow{ 2}{*}{`nsubj\_amod'} & (new \textit{subject}, fly) (5.50/10)\\
				& (new pilot, Capable Of, fly solo)\\
				
				\bottomrule
			\end{tabular}
		}
		\subtable[\normalsize Unusual group in SP-10K]{
			\begin{tabular}{c|c}
				\toprule 
				SP relation & SP pair versus OMCS triplets \\
				\midrule
				\multirow{ 2}{*}{`dobj'} & (buy, health) (2.00/10)\\
				& (health food store, UsedFor, buy vitamin)\\
				\midrule
				\multirow{ 2}{*}{`nsubj'} &(line, run) (2.00/10)\\
				& (leave line, UsedFor, run to restroom) \\
				\midrule
				\multirow{ 2}{*}{`amod'} & (outside, dog) (3.50/10)\\
				& (small dog, AtLocation, outside)\\
				\midrule
				\multirow{ 2}{*}{`dobj\_amod'} & (enjoy, hungry \textit{object}) (2.00/10)\\
				& (enjoy day, HasSubevent, get hungry)\\
				\midrule
				\multirow{ 2}{*}{`nsubj\_amod'} & (wet \textit{subject}, become) (3.25/10)\\
				& (get wet, Causes, become cold)\\
				
				\bottomrule
			\end{tabular}
		}
		\subtable[\normalsize Impossible group in SP-10K]{
			\begin{tabular}{c|c}
				\toprule 
				SP relation & SP pair versus OMCS triplets \\
				\midrule
				\multirow{ 2}{*}{`dobj'} & (eat, mail) (0.00/10)\\
				& (mail letter, HasSubevent, eat cheese)\\
				\midrule
				\multirow{ 2}{*}{`nsubj'} & (set, sleep) (1.00/10)\\
				& (sleep, HasSubevent, set an alarm)\\
				\midrule
				\multirow{ 2}{*}{`amod'} & (red, child) (0.68/10)\\
				& (child wagon, HasProperty, red)\\
				\midrule
				\multirow{ 2}{*}{`dobj\_amod'} & (drive, bottom \textit{object}) (1.50/10)\\
				& (drive car, HasSubevent, bottom out)\\
				\midrule
				\multirow{ 2}{*}{`nsubj\_amod'} & (fun \textit{subject}, hurt) (1.50/10)\\
				& (having fun, HasSubevent, get hurt)\\
				\bottomrule
			\end{tabular}
		}
		\caption{\normalsize Examples of covered SP pairs and corresponding OMCS triplets (Partial Match).}
	\end{table*}
		\clearpage

\end{document}